\documentclass[10pt,twocolumn,letterpaper]{article}

\usepackage{cvpr}              

\usepackage{colortbl}%
\usepackage{graphicx}
\usepackage{amsmath}
\usepackage{amssymb}
\usepackage{booktabs,caption}
\usepackage{multirow}   
\usepackage{bm}
\usepackage{dsfont}
\usepackage[linesnumbered,ruled,vlined]{algorithm2e}
\usepackage{makecell}
\usepackage[flushleft]{threeparttable}
\usepackage{enumitem}
\usepackage{siunitx}

\definecolor{MyBlack}{HTML}{323A45}

\definecolor{cvprblue}{rgb}{0.21,0.49,0.74}
\usepackage[pagebackref,breaklinks,colorlinks,citecolor=cvprblue]{hyperref}
\usepackage[capitalize]{cleveref}

\def\paperID{52} 
\def\confName{CVPR}
\def\confYear{2025}

\begin{document}
\title{AIM 2025 challenge on Inverse Tone Mapping Report: Methods and Results}

\author{
Chao Wang$^*$\and
Francesco Banterle $^*$\and
Bin Ren$^*$\and
Radu Timofte$^*$ \and
Xin Lu \and 
Yufeng Peng \and
Chengjie Ge \and
Zhijing Sun \and
Ziang Zhou \and
Zihao Li \and
Zishun Liao \and
Qiyu Kang \and
Xueyang Fu \and
Zheng-Jun Zha \and
Zhijing Sun \and
Xingbo Wang \and
Kean Liu \and
Senyan Xu \and
Yang Qiu \and
Yifan Ding \and
Gabriel Eilertsen \and
Jonas Unger \and
Zihao Wang \and
Ke Wu \and
Jinshan Pan \and
Zhen Liu \and
Zhongyang Li \and
Shuaicheng Liu \and
S.M Nadim Uddin
}
\maketitle
\let\thefootnote\relax\footnotetext{
$^*$ 
C. Wang (winchao1984@gmail.com, MPI-Info, Germany \& Peng Cheng Laboratory, China), 
F. Banterle (francesco.banterle@isti.cnr.it, ISTI-CNR, Italy), 
B. Ren (bin.ren@unitn.it, University of Pisa, \&  University of Trento, Italy), 
and R. Timofte (Radu.Timofte@uni-wuerzburg.de,
University of W\"urzburg, Germany)
were the challenge organizers, while the other authors participated in the challenge.\\ 
Appendix~\ref{sec:teams} contains the authors' teams and affiliations.\\
AIM 2025 webpage: \url{https://cvlai.net/aim/2025/}.
}

\begin{abstract}
    This paper presents a comprehensive review of the AIM 2025 Challenge on Inverse Tone Mapping (ITM). The challenge aimed to push forward the development of effective ITM algorithms for HDR image reconstruction from single LDR inputs, focusing on perceptual fidelity and numerical consistency. A total of \textbf{67} participants submitted \textbf{319} valid results, from which the best five teams were selected for detailed analysis. This report consolidates their methodologies and performance, with the lowest PU21-PSNR among the top entries reaching 29.22 dB. The analysis highlights innovative strategies for enhancing HDR reconstruction quality and establishes strong benchmarks to guide future research in inverse tone mapping.
\end{abstract}

\section{Introduction}
\label{sec:introduction}

Inverse Tone Mapping (ITM) aims to reconstruct high dynamic range (HDR) images from single low dynamic range (LDR) inputs. This process involves several key steps, including de-quantization, de-contouring, linearization, dynamic range expansion, completion of overexposed and underexposed regions, and color gamut extension~\cite{banterle2017advanced}. 
Inverse Tone Mapping (ITM) is essential for displaying legacy LDR content on HDR devices, enabling existing images and videos to benefit from the wider dynamic range and color gamut of modern displays. With the rapid adoption of HDR technology, ITM has gained importance not only in consumer entertainment but also in applications such as digital archiving, photography, gaming, and VR/AR~\cite{banterle2022unsupervised,wang2025lediff,li2025scenesplat}.

Before the rise of deep learning, ITM methods were primarily model-driven~\cite{meylan2006reproduction,meylan2007tone,banterle2006inverse,akyuz2007hdr,banterle2007framework,rempel2007ldr2hdr,banterle2008expanding,ren2025tenth,didyk2008enhancement,kovaleski2009high,masia2017dynamic}. These approaches focused on linearization and dynamic range expansion using analytical tone-mapping functions or inverse camera response models. While they perform adequately for high-quality LDR inputs with minimal saturation or quantization artifacts, they fail to reconstruct missing information in heavily overexposed or underexposed regions.

With the advent of deep learning, ITM has significantly advanced through data-driven models that can restore lost details and hallucinate plausible content in saturated areas. Existing learning-based methods can be broadly divided into direct and indirect approaches. Direct methods reconstruct HDR images from LDR inputs in an end-to-end manner. HDRCNN~\cite{EKDMU17} pioneered this direction with a VGG-based~\cite{simonyan2014very} CNN and a luminance-reflectance decomposition loss to improve reconstruction quality. \cite{marnerides2018expandnet} introduced ExpandNet, a multi-branch network capturing local, medium, and global context to mitigate U-Net artifacts. MaskHDR\cite{santos2020single} employed feature masks inspired by partial convolutions~\cite{liu2018image} to emphasize unsaturated regions, while SingleHDR~\cite{liu2020single} decomposed ITM into subproblems such as de-quantization and linearization, using specialized sub-networks for each. \cite{yu2021luminance} designed a two-stream architecture balancing saturated and unsaturated regions, and \cite{dille2025intrinsic} adopted intrinsic decomposition to separately reconstruct shading and albedo, ensuring color fidelity.

Indirect methods first synthesize virtual exposure stacks from the input and then merge them into HDR. DrTMO~\cite{endo2017deep} combined a 2D U-Net encoder with a 3D U-Net~\cite{cciccek20163d} decoder to generate exposure sequences, followed by merging based on~\cite{debevec2023recovering}.  \cite{lee2018deep} adapted the EDSR architecture~\cite{lim2017enhanced} for cascaded exposure generation, later enhanced by recursive designs to reduce parameters~\cite{lee2018deep1}. \cite{le2023single} enforced feature consistency across exposures using representation loss, and~\cite{zhang2023revisiting} introduced Swin-Transformer~\cite{liu2021swin,ren2024sharing,ren2023masked} for long-range dependency modeling with an end-to-end fusion.

Beyond static images, ITM has been extended to video. ~\cite{kim2019deep} synthesized training data from HDR videos and proposed a two-branch network that jointly addresses ITM and super-resolution by separating low and high frequency components, further improved with GAN regularization~\cite{kim2020jsi}. \cite{he2022sdrtv} designed a context-aware hierarchical module for modeling global-local relationships, enhancing temporal consistency. \cite{kalantari2019deep} treated adjacent frames as implicit multi-exposure inputs, combining them through a pyramid-based optical flow alignment. \cite{banterle2022unsupervised} further advanced this by exploiting inter-frame similarity to transfer details from unsaturated frames to saturated ones without supervision.

Recently, generative models have also been introduced into ITM, opening new possibilities for detail synthesis and enhanced visual realism. These approaches leverage generative priors to hallucinate missing content and improve reconstruction quality, representing a promising direction for future research.
GlowGAN~\cite{wang2023glowgan} learns HDR generation from unlabeled LDR data via a differentiable camera model and employs GAN inversion ~\cite{shmelkov2018good,ren2021cascaded,tang2025hierarchical,xia2022gan} for ITM, reducing dependency on HDR ground truth. Building on this concept, \cite{bemana2024exposure} and \cite{goswami2024semantic} extend it to diffusion models, enabling multi-exposure synthesis and HDR reconstruction without fine-tuning, marking a paradigm shift toward generative modeling for ITM. Moreover, LEDiff~\cite{wang2025lediff} finetunes a pretrained Stable Diffusion model~\cite{rombach2022high} on a small HDR dataset, greatly improving the reconstruction of overexposed and underexposed regions and enabling photorealistic HDR synthesis.

As HDR displays become increasingly prevalent and the demand for high-quality visual content rises, ITM plays a critical role in bridging the gap between existing LDR media and emerging HDR standards. Efficient and accurate ITM solutions not only enable improved visual experiences but also set the foundation for advanced applications such as HDR video streaming, immersive media, and content remastering, but also ensure long-term compatibility and accessibility of visual content across diverse devices and display technologies.

In collaboration with the Advances in Image Manipulation (AIM 2025) workshop, we organize the AIM 2025 Challenge on Inverse Tone Mapping. The goal is to reconstruct high-quality HDR images from single-exposure LDR inputs, with emphasis on faithful dynamic range expansion and detail recovery. Participants optimize for PU21-PSNR and PU21-SSIM to ensure both radiometric accuracy and perceptual quality. The challenge promotes practical and innovative solutions, establishes standardized evaluation protocols, and advances the development of robust HDR reconstruction systems. For context, the lowest scores achieved on the development and test sets are 29.83 / 29.22 for PU21-PSNR and 0.85 / 0.85 for PU21-SSIM, respectively.


This challenge is one of the AIM 2025~\footnote{\url{https://www.cvlai.net/aim/2025/}} workshop associated challenges on: high FPS non-uniform motion deblurring~\cite{aim2025highfps}, rip current segmentation~\cite{aim2025ripseg}, inverse tone mapping~\cite{aim2025tone}, robust offline video super-resolution~\cite{aim2025videoSR}, low-light raw video denoising~\cite{aim2025videodenoising}, screen-content video quality assessment~\cite{aim2025scvqa}, real-world raw denoising~\cite{aim2025rawdenoising}, perceptual image super-resolution~\cite{aim2025perceptual}, efficient real-world deblurring~\cite{aim2025efficientdeblurring}, 4K super-resolution on mobile NPUs~\cite{aim20254ksr}, efficient denoising on smartphone GPUs~\cite{aim2025efficientdenoising},  efficient learned ISP on mobile GPUs~\cite{aim2025efficientISP}, and stable diffusion for on-device inference~\cite{aim2025sd}. Descriptions of the datasets, methods, and results can be found in the corresponding challenge reports.

\section{AIM 2025 Inverse Tone Mapping Challenge}

\subsection{Overview}
The primary targets of the challenge are:
\begin{itemize}
    \item Promoting research in the area of single-image inverse tone mapping for HDR reconstruction.
    \item Facilitating fair and comprehensive comparisons between the efficiency and accuracy of various ITM methods.
    \item Providing a platform for academic and industrial participants to engage, exchange ideas, and foster collaborations in advancing HDR technologies.
\end{itemize}

Specifically, the competition addresses four key tasks:
\begin{enumerate}[label=(\roman*)]
    \item \textbf{Compression artifact removal}: Many LDR images are stored in compressed formats such as JPEG. This introduces blocking and similar artifacts that require elimination.
    \item \textbf{Denoising}: Noise often appears in dark regions of LDR images due to sensor limitations. Effective denoising is therefore essential.
    \item \textbf{Linearization and dynamic range expansion}: LDR images typically use nonlinear compression within a limited range. HDR reconstruction demands reversing this process and expanding the range.
    \item \textbf{Completion of over/under exposed regions}: Overexposed highlights and underexposed shadows result in missing information. Restoring these regions is crucial for high-quality HDR output.
\end{enumerate}

\begin{figure*}[!t]
    \centering
    \includegraphics[width=\linewidth]{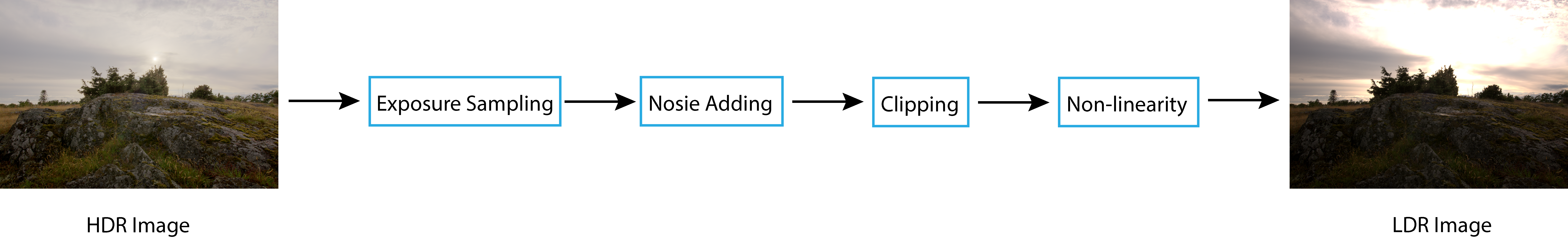}  
    \caption{ITM Dataset creation pipeline.}
    \label{fig:ITM_data_pipeline}
\end{figure*}

\subsection{Dataset}
\label{sec:dataset}
We collect HDR images from several publicly available datasets~\cite{froehlich2014creating,tel2023alignment,hanji2022sihdr,liu2020single,panetta2021tmo} and evaluate performance on the benchmark from~\cite{fairchild2007hdr}, and on additional unpublished samples. Since these HDR images are stored in relative luminance values, we apply normalization prior to processing, following the recommendation in~\cite{wang2025lediff}. To synthesize LDR data, we first estimate the maximum valid exposure range of each HDR image using the method described in~\cite{andersson2021visualizing}, then sample exposures within this range and inject noise into the sampled images. Pixel values are subsequently clipped to the range [0, 1]. Following~\cite{EKDMU17}, we sample a camera response function from clustered CRF databases and apply it to achieve nonlinearity, thereby simulating the behavior of a real imaging pipeline. Finally, the processed images are stored as 8-bit JPEG LDR images. The overall dataset preparation pipeline is illustrated in Figure~\ref{fig:ITM_data_pipeline}.

\subsection{Challenge Description}

The objective of this challenge is to reconstruct high dynamic range (HDR) images from single low dynamic range (LDR) inputs. The dataset consists of paired LDR-HDR images generated from real HDR content through tone mapping, exposure sampling, and noise simulation. The training set contains approximately 19{,}000 LDR-HDR pairs at a resolution of 256$\times$256, while the validation and test sets include 100 images each at 512$\times$512 resolution.

\medskip
\noindent\textbf{Challenge phases:}

\textit{(1) Development and validation phase}: Participants are provided with the full training set and the validation set. The validation data serve for online evaluation, where participants upload HDR reconstructions to the evaluation server and receive PU21-PSNR and PU21-SSIM scores. Baseline code and evaluation scripts are available to assist in model development and ensure reproducibility.

\textit{(2) Testing phase}: During the final test phase, participants receive only the 100 LDR test images, with the corresponding HDR ground truths kept hidden for unbiased evaluation. Participants must submit their reconstructed HDR outputs to the evaluation server and provide both their code and a factsheet. The organizers verify and execute the submitted code to compute final scores, which are shared with participants after the challenge concludes.

\medskip
\noindent\textbf{Evaluation Metrics:}  
Performance is evaluated via perceptually uniform metrics specifically designed for HDR image quality assessment, namely PU21-PSNR and PU21-SSIM, as suggested in~\cite{azimi2021pu21}. To reduce dataset bias, we apply camera response curve correction following the approach in~\cite{hanji2022sihdr}.

In addition, participants are encouraged to report the number of parameters and inference runtime. 
A U-Net-based baseline model is provided for reference, achieving PU21-PSNR of 27.23 and PU21-SSIM of 0.91 on the validation set.

\begin{table}[!t]
    \centering
    \caption{PU21-PSNR and PU21-SSIM results of the challenge on the test set.}
    \begin{tabular}{lcc}
    \hline
    \textbf{Team} & \textbf{PU21-PSNR (dB)} & \textbf{PU21-SSIM} \\
    \midrule
    ToneMapper & \textbf{34.49} & \textbf{0.95} \\
    HDRer & 34.39 & 0.95 \\
    LiU\_CGIP & 34.33 & 0.95 \\
    UESTC-ITM & 34.06 & 0.94 \\
    Jowgik (DITM) & 33.64 & 0.94 \\
    NJ Challenger & 29.22 & 0.85 \\
    \hline
    \end{tabular}
    \label{TAB:RESULTS:FINAL}
\end{table}

\section{Challenge Results}
\label{sec:esr_results}
The AIM 2025 Challenge on Inverse Tone Mapping attracted 69 participants and recorded a total of 319 submissions. Six teams submitted final results. All entries were evaluated based on PU21-PSNR and PU21-SSIM, which reflect both radiometric accuracy and perceptual quality in HDR reconstruction, see Table~\ref{TAB:RESULTS:FINAL}. The best performance on the test set was achieved by \textbf{ToneMapper}, reaching a PU21-PSNR of 34.49 dB and PU21-SSIM of 0.95. \textbf{HDRer} and \textbf{LiU\_CGIP} closely followed with PSNRs of 34.39 dB and 34.33 dB, respectively, both also reaching an SSIM of 0.95. At the lower end, \textbf{NJ Challenger} recorded the minimum scores with 29.22 dB in PSNR and 0.85 in SSIM, demonstrating the difficulty of restoring severely saturated or underexposed regions. Across all teams, the PU21-PSNR scores ranged from 29.83 to 34.58 dB on the development set and 29.22 to 34.49 dB on the test set; PU21-SSIM ranged from 0.85 to 0.95 on both sets, indicating relatively tighter variance in perceptual structure despite broader fidelity differences.

\textbf{ToneMapper} adopts a NAFNet-based architecture designed for efficient multi-scale image restoration. Its key innovation lies in the use of regularization-enhanced training, inspired by EEDTP, which improves generalization under diverse LDR degradations. The team employed a three-stage training scheme with progressively increasing patch sizes and decreasing learning rates, trained entirely on the official dataset. The method balances performance and efficiency, with moderate complexity (27.18M parameters, 4.84 GFLOPs) and high-quality output.

\textbf{HDRer} builds on the same NAFNet backbone but introduces diffusion-based data augmentation to diversify training inputs. Instead of architectural changes, the method generates additional synthetic LDR scenes to mimic varied dynamic range conditions. These are then incorporated into the same three-stage training pipeline as ToneMapper. The approach reduces model size (15.51M parameters) while achieving nearly equivalent performance, demonstrating the effectiveness of data-centric strategies for ITM.

\textbf{LiU\_CGIP} proposes a diffusion-based generative model using the Refusion framework. It formulates inverse tone mapping as a score-based restoration problem in the PU21 space. A modified NAFNet backbone estimates score functions across time steps within a stochastic differential equation framework. Extensive external HDR data (~4,000 images) is used for training, with additional SSIM-regularized loss terms improving perceptual consistency. Although computationally demanding (131.4M parameters, 253.6 GFLOPs, ~6s per image), the model achieves near state-of-the-art performance and provides valuable insights into the challenges of restoring saturated regions.

\textbf{NJ Challenger} presents a modular approach that reverses the camera imaging pipeline. It decomposes ITM into 3 stages: dequantization, linearization, and hallucination. Each stage is handled by a dedicated network (U-Net, ResNet18, and VGG-based encoder-decoder), trained separately with specialized loss functions. The method incorporates prior knowledge such as CRF constraints and perceptual losses from pretrained VGG features. While conceptually well-grounded, the method underperforms in both PSNR and SSIM, highlighting the challenge of modeling each stage independently in complex HDR reconstructions.

\textbf{UESTC-ITM} proposes a dual-branch framework called ITMFlow that fuses generative modeling and deterministic regression. The first branch uses conditional flow matching to reconstruct PU21-encoded HDR from noisy LDR inputs, while the second branch employs a hybrid CNN+ViT architecture optimized for PU metrics. Both branches are trained independently and fused at test time via averaging. Despite using only the official dataset, the method achieves high quantitative and perceptual performance (34.06 dB PSNR, 0.94 SSIM) with manageable complexity (69.79M parameters, ~1.4s per image), demonstrating the effectiveness of hybrid architectures.

\textbf{Jowgik (DITM)} introduces a progressive pipeline comprising BitRecoverNet, BitExpansionNet, and DetailEnhancementNet. The model first recovers quantized bits and decomposes the image into exposure-based components before gradually expanding the dynamic range and refining details. Multiple loss terms, including $\mu$-law compressed L1, PU-domain SSIM, and patch-wise histogram matching learning at various stages. With only 1.97M parameters and 25.64 GFLOPs, the model offers an efficient solution with strong visual quality and competitive test scores (33.64 dB PSNR, 0.94 SSIM), validating the potential of low-complexity modular pipelines for HDR restoration.

\section*{Acknowledgments}
This work was partially supported by the Alexander von Humboldt Foundation. We thank the AIM 2025 sponsors: AI Witchlabs and University of W\"urzburg (Computer Vision Lab).

\section{Challenge Methods and Teams}
\label{sec:methods_and_teams}
\subsection{ToneMapper: Boosting Inverse Tone Mapping via Regularization Training}

\begin{figure}[!t]
    \centering
    \includegraphics[width=0.99\linewidth]{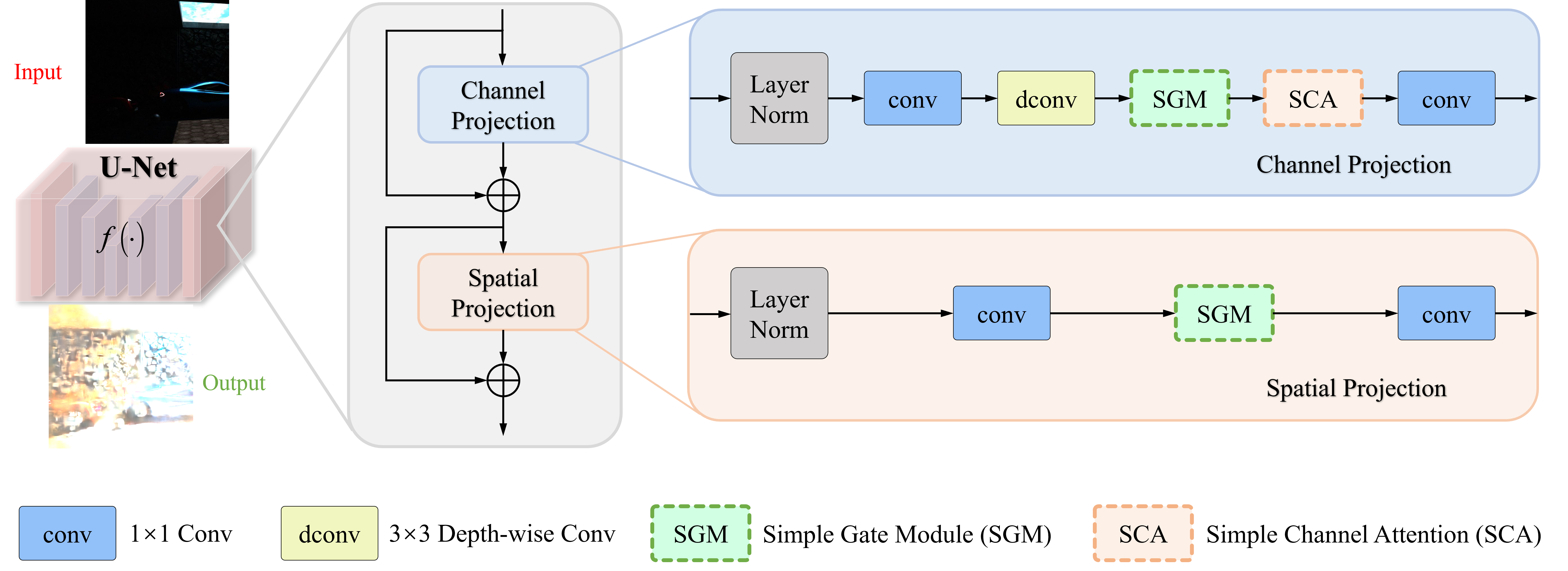}
    \caption{\textit{ToneMapper}'s team framework.}
    \label{team01:framework}
\end{figure}

\noindent \textbf{General method description}. The proposed network architecture is mainly based on NAFNet~\cite{Chen2022SimpleBF}, which fuses multi-scale image information through a series of convolutional layers and skip connections to process LDR images into HDR images~\cite{Lu_2025_CVPR_AALN,Lu_2024_CVPR_HirFormer,Lu_2025_CVPR_ILAWR,Lu_2025_CVPR_EvenFormer}. It should be noted that inspired by EEDTP~\cite{lu2025elucidating}, we use regularized enhancement training to improve the performance of ordinary restoration models, which greatly improves the image processing effect. Please see Figure \ref{team01:framework}.               

\noindent \textbf{Total method complexity}.
\begin{itemize}
\item Number of parameters = 27.18M
\item FLOPs = 4.84G
\item GPU memory consumption: A GTX1080 is sufficient
\end{itemize}

\noindent\textbf{Training strategy}. Inverse Tone Mapping aims to reconstruct a high dynamic range from a low dynamic range input image. Due to the complexity of dynamic scenes, LDR images with different exposures have the same complex degradation conditions, which poses a severe test for the generalization of the image restoration network. Therefore, we used the generalization-enhanced regularization method in EEDTP~\cite{lu2025elucidating} to promote better restoration of HDR effects.

Our training process is divided into three stages:
\begin{enumerate}
    \item  We adopt the Adam optimizer with a batch size of 100 and the patch size of $64\times64$. The initial learning rate is $4\times 10^{-4}$ and changes with Cosine Annealing scheme, including 1000 epochs in total. The first stage is performed on the NVIDIA 4090D device. We obtain the best model at this stage as the initialization of the second stage.

    \item We adopt the Adam optimizer with a batch size of 50 and the patch size of $128 \times 128$. The initial learning rate is $4\times 10^{-5}$ and changes with Cosine Annealing scheme, including 300 epochs in total. The second stage is performed on the NVIDIA 4090D device and uses gradient accumulation. We obtain the best model at this stage as the initialization of the next stage.
 
    \item We adopt the SGD optimizer with a batch size of 22 and the patch size of $256 \times 256$. The initial learning rate is $2\times 10^{-5}$ and changes with Cosine Annealing scheme, including 200 epochs in total. The third stage is performed on the NVIDIA 4090D device and uses gradient accumulation.
\end{enumerate}

\noindent\textbf{Testing strategy}. During the test time, we adopt the model after fine-tuning to get the best performance. Moreover, we utilize input-ensemble strategy to obtain the best results. We test the model on NVIDIA 4090D.

\noindent\textit{Results of the comparison to other approaches}.
We tested the previous SOTA models HirFormer \cite{Lu_2024_CVPR_HirFormer}, EvenFormer \cite{Lu_2025_CVPR_EvenFormer}, AALN \cite{Lu_2025_CVPR_AALN}, ILAWR \cite{Lu_2025_CVPR_ILAWR}, and EEDTP \cite{lu2025elucidating} in motion deblurring. HirFormer and EvenFormer have low training efficiency due to high video memory usage, while AALN and ILAWR cannot achieve optimal performance.

\noindent\textit{Experimental results}. In quantitative results, our approach achieves the highest PSNR on the AIM 2025 Inverse Tone Mapping Challenge test dataset.

\noindent\textbf{Best scores for development/testing}: 
\begin{itemize}
    \item PU21-PSNR: 34.58 / 34.49 	
    \item PU21-SSIM: 0.95 / 0.95
\end{itemize}

\subsection{HDRer: Deep High Dynamic Range Imaging via Dynamic Scenes Generation}

\begin{figure}[!t]
    \centering
    \includegraphics[width=0.88\linewidth]{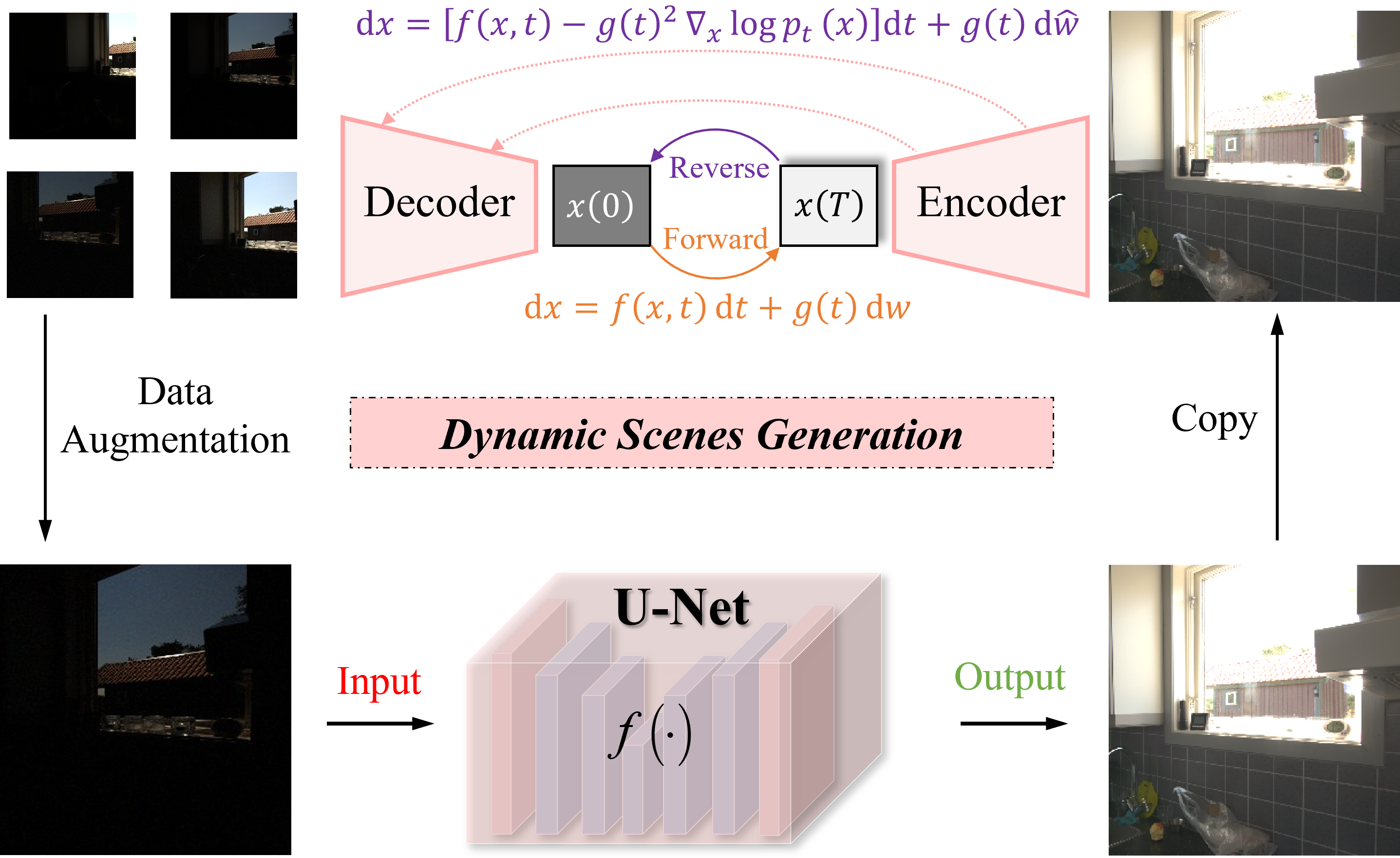}
    \caption{\textit{HDRer}'s team dynamic scenes generation.}
    \label{team02:Scenes}
\end{figure}

\begin{figure}[!t]
    \centering
    \includegraphics[width=0.99\linewidth]{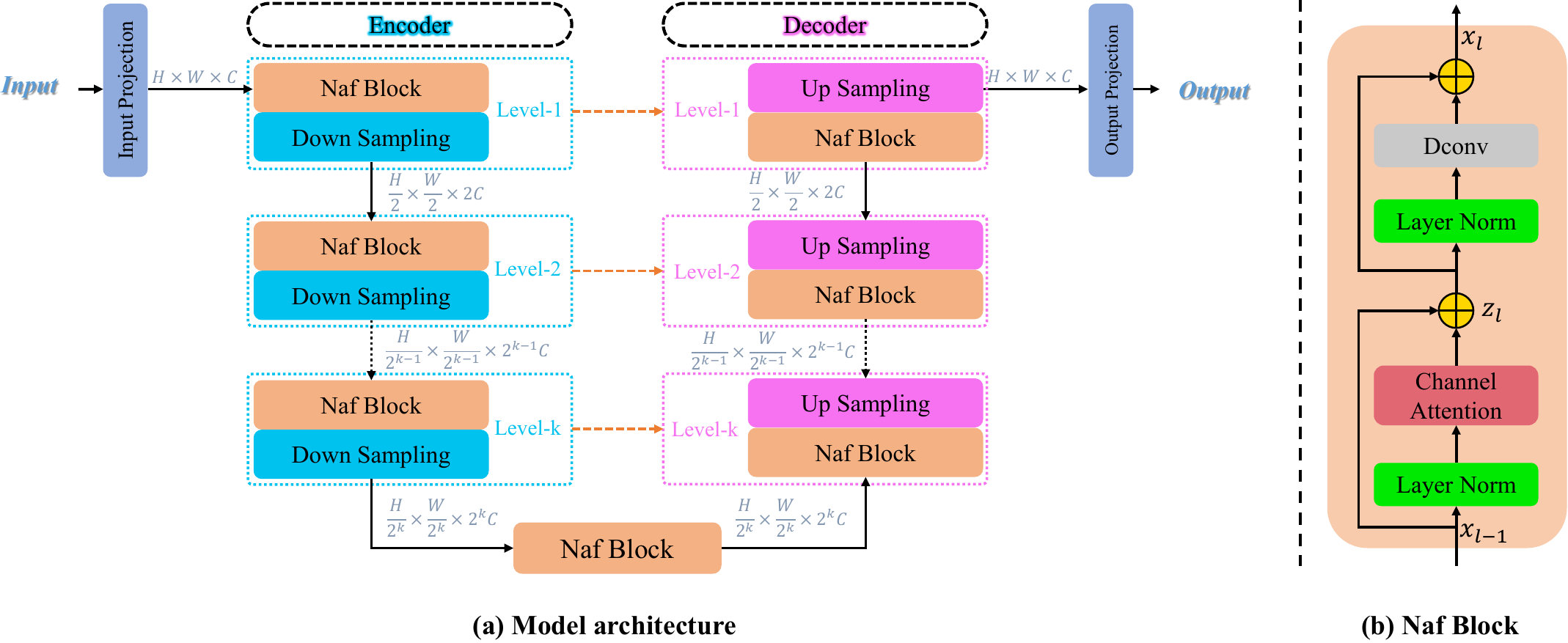}
    \caption{\textit{HDRer}'s team framework.}
    \label{team02:framework}
\end{figure}

\noindent\textbf{General method description}. The proposed network architecture is mainly based on NAFNet~\cite{Chen2022SimpleBF}, which fuses multi-scale image information through a series of convolutional layers and skip connections to process LDR images into HDR images~\cite{Lu_2025_CVPR_AALN,Lu_2024_CVPR_HirFormer,Lu_2025_CVPR_ILAWR,Lu_2025_CVPR_EvenFormer}. It should be noted that inspired by AALN~\cite{Lu_2025_CVPR_AALN}, we use Diffusion to generate more low-dynamic scene images to enhancement training to improve the performance of ordinary restoration models, which greatly improves the image processing effect. Please see Figure \ref{team02:framework} and Figure \ref{team02:Scenes}.

\noindent \textbf{Total method complexity}.
\begin{itemize}
\item Number of parameters = 15.51M
\item FLOPs = 2.80G
\item GPU memory consumption: A GTX1080 is sufficient
\end{itemize}

\noindent\textbf{Training strategy}. Our training process is divided into three stages:

\begin{enumerate}

\item We adopt the Adam optimizer with a batch size of 100 and the patch size of $64\times64$. The initial learning rate is $4\times 10^{-4}$ 
and changes with Cosine Annealing scheme, including 1000 epochs in total. The first stage is performed
on the NVIDIA 4090D device. We obtain the best model at this stage as the initialization of the second stage.

\item We adopt the Adam optimizer with a batch size of 50 and the patch size of $128 \times 128$. The initial learning rate is $4\times 10^{-5}$ and changes with Cosine Annealing scheme, including 300 epochs in total. The second stage is performed on the NVIDIA 4090D device and uses gradient accumulation. We obtain the best model at this stage as the initialization of the next stage.
 
\item We adopt the SGD optimizer with a batch size of 22 and the patch size of $256 \times 256$. The initial learning rate is $2\times 10^{-5}$ and changes with Cosine Annealing scheme, including 200 epochs in total. The third stage is performed on the NVIDIA 4090D device and uses gradient accumulation.
\end{enumerate}

\noindent\textbf{Testing strategy}. During the test time, we adopt the model after fine-tuning to get the best performance. Moreover, we utilize input-ensemble strategy to obtain the best results. We test the model on NVIDIA 4090D.

\noindent\textit{Results of the comparisons to other approaches results}. We tested the previous SOTA models HirFormer \cite{Lu_2024_CVPR_HirFormer}, EvenFormer \cite{Lu_2025_CVPR_EvenFormer}, AALN \cite{Lu_2025_CVPR_AALN}, ILAWR \cite{Lu_2025_CVPR_ILAWR}, and EEDTP \cite{lu2025elucidating} in motion deblurring. HirFormer and EvenFormer have low training efficiency due to high video memory usage, while AALN and ILAWR cannot achieve optimal performance.

\noindent\textit{Experimental results}. In quantitative results, our approach achieves the 2nd highest PSNR on the AIM 2025 Inverse Tone Mapping Challenge test dataset.

\noindent\textbf{Best scores for development/testing}: 
\begin{itemize}
    \item PU21-PSNR: 33.09 / 34.39	
    \item PU21-SSIM: 0.93 / 0.95
\end{itemize}

\subsection{LiU\_CGIP: What Makes Inverse Tone Mapping Hard? An Empirical Study of Error Patterns and Perceptual Challenges}

\begin{figure}[!t]
    \centering
    \includegraphics[width=\linewidth]{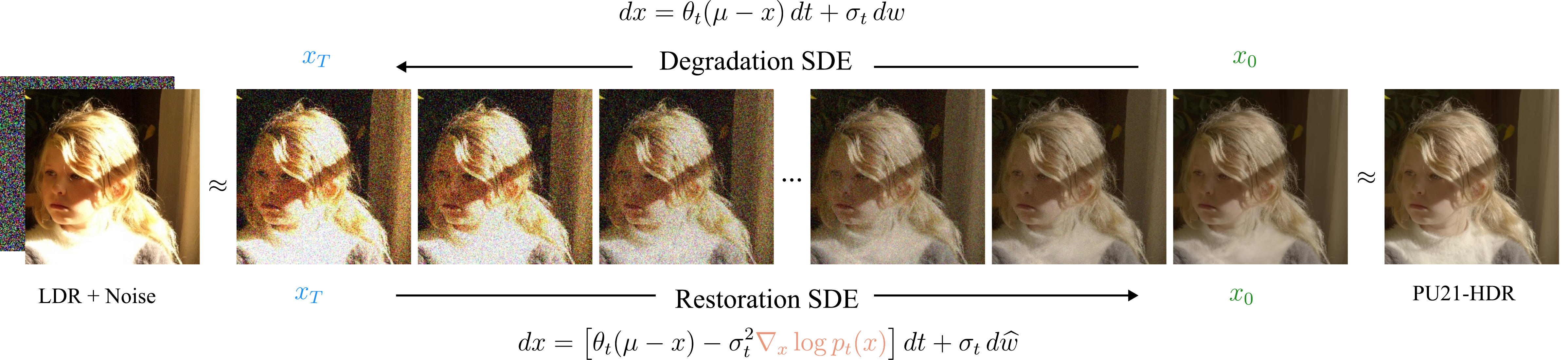}
    \caption{Pipeline of \textit{LiU\_CGIP}'s solution: The LDR input is processed through the restoration SDE to generate a PU21-HDR representation, which is subsequently decoded to produce the final HDR output.}
    \label{team03:fig:pipeline}
\end{figure}

\noindent\textbf{General method description}. We employ Refusion~\cite{luo2023refusion}, a score-based diffusion model as the backbone of our solution for the AIM 2025 ITM Challenge. Refusion is characterized by two key components. First, it adopts mean-reverting stochastic differential equations (SDE)~\cite{luo2023image}, a specialized formulation of the diffusion SDE~\cite{song2021scorebased}, which requires both the noise and the degraded image as inputs. Second, instead of the commonly used U-Net architecture~\cite{ronneberger2015u}, Refusion utilizes a modified NAFBlock \cite{chen2022simple} as its backbone, offering improved performance and less computational cost in image reconstruction tasks. As illustrated in Figure~\ref{team03:fig:pipeline}, we first map HDR images to the PU21 space~\cite{azimi2021pu21}, then add noise to their corresponding LDR counterparts in order to reconstruct PU21-HDR. Finally, these PU21-HDR representations are decoded back to HDR images. Similar to other score-based diffusion models, Refusion employs a forward degradation process governed by an SDE:
\begin{equation}
dx = \theta_t (\mu - x) \, dt + \sigma_t \, dw.
\label{eqn:f-sde}
\end{equation}
Given this forward SDE, the corresponding backward restoration SDE is formulated as
\begin{equation}
dx = \left[ \theta_t (\mu - x) - \sigma_t^2 \nabla_x \log p_t(x) \right] dt + \sigma_t \, d\hat{w}.
\label{eqn:b-sde}
\end{equation}
To train the model, Refusion minimizes the following objective function
\begin{equation}
J_\gamma(\phi) := \sum_{i=1}^{T} \gamma_i \, \mathbb{E} [ \|  x_{i-1} - x_{i-1}^* \| ],
\end{equation}
\noindent where \( x_{i-1}^* \) is derived by \cite{luo2023refusion} via maximum likelihood estimation, and \( x_{i-1} \) is obtained from Eq.~(\ref{eqn:b-sde}). The score function \( \nabla_x \log p_t(x) \) is approximated using a modified NAFNet architecture parameterized by \( \phi \). The weights \( \gamma_1, \dots, \gamma_T \) are positive scalars that balance the contributions of each timestep. For further details, we refer the reader to \cite{luo2023refusion}. However, empirical results show that using the \( \ell_1 \) loss yields better performance than the \( \ell_2 \) loss. Therefore, we incorporate an additional SSIM-based regularization term to formulate the total score-matching loss as:
\begin{equation}
J_\gamma(\phi) := \sum_{i=1}^{T} \gamma_i \, \mathbb{E} [ | x_{i-1} - x_{i-1}^* | + \lambda \cdot \ell_{\mathrm{SSIM}}(x_{i-1}, x_{i-1}^*) ],
\end{equation}
\noindent where \(\ell_{\mathrm{SSIM}}(x, \hat{x}) =  1 - \mathrm{SSIM}(x, \hat{x})\) and \(\lambda\) is a positive weight.

\noindent\textbf{Total method complexity}. The final model, Refusion, contains 131.4M parameters. For a \(512 \times 512\) input, the computational cost is 253.6G FLOPs, with an average runtime of approximately 6 seconds on a single A100 GPU.

\noindent\textbf{Training strategy}. We utilize additional data from the HDRCNN dataset~\cite{EKDMU17}. Detailed statistics are provided in Table~\ref{team03:tab:data}. For generating the corresponding LDR images, we use HDRCNN's virtual camera simulation\footnote{Virtual camera code: \href{https://github.com/gabrieleilertsen/hdrcnn}{https://github.com/gabrieleilertsen/hdrcnn}} to synthesize 104K image pairs at a resolution of \(256 \times 256\). The simulation includes random cropping, camera response curve mapping, exposure clipping, and Gaussian noise injection, and then normalization to [0, 1000]. Further implementation details are available in Appendix A of HDRCNN~\cite{EKDMU17}.

We first train the model solely on the HDRCNN dataset~\cite{EKDMU17}. The training uses the LION optimizer~\cite{chen2023symbolic} with an initial learning rate of \(4 \times 10^{-5}\). Cosine annealing is applied without warm-up. The model is trained for 700{,}000 iterations with a batch size of 8, taking approximately 2.5 days on a single NVIDIA A100 GPU. We then fine-tune the pretrained model jointly on both the HDRCNN dataset and the official AIM Inverse Tone Mapping Challenge dataset for another 700{,}000 iterations, using a reduced learning rate of \(1 \times 10^{-5}\). All models are trained from scratch without using any external pretrained weights. For the diffusion process, we use a cosine noise schedule with 100 discretized time steps.

\noindent\textbf{Testing strategy}. We do not employ any test-time augmentation. During inference, we simply run the backward restoration SDE to reconstruct PU21-HDR images, which are then decoded to linear HDR. 

\noindent\textit{Quantitative Results}. Our qualitative results on the test set are summarized in Table~\ref{team03:tab:pu21-comparison}. Training with additional data significantly improves both PU21-PSNR and PU21-SSIM metrics. Note that we did not submit other model variants to the test leaderboard, therefore, further ablation studies would require access to ground truth data for the test set.

\begin{figure}[!t]
    \centering
    \includegraphics[width=\linewidth]{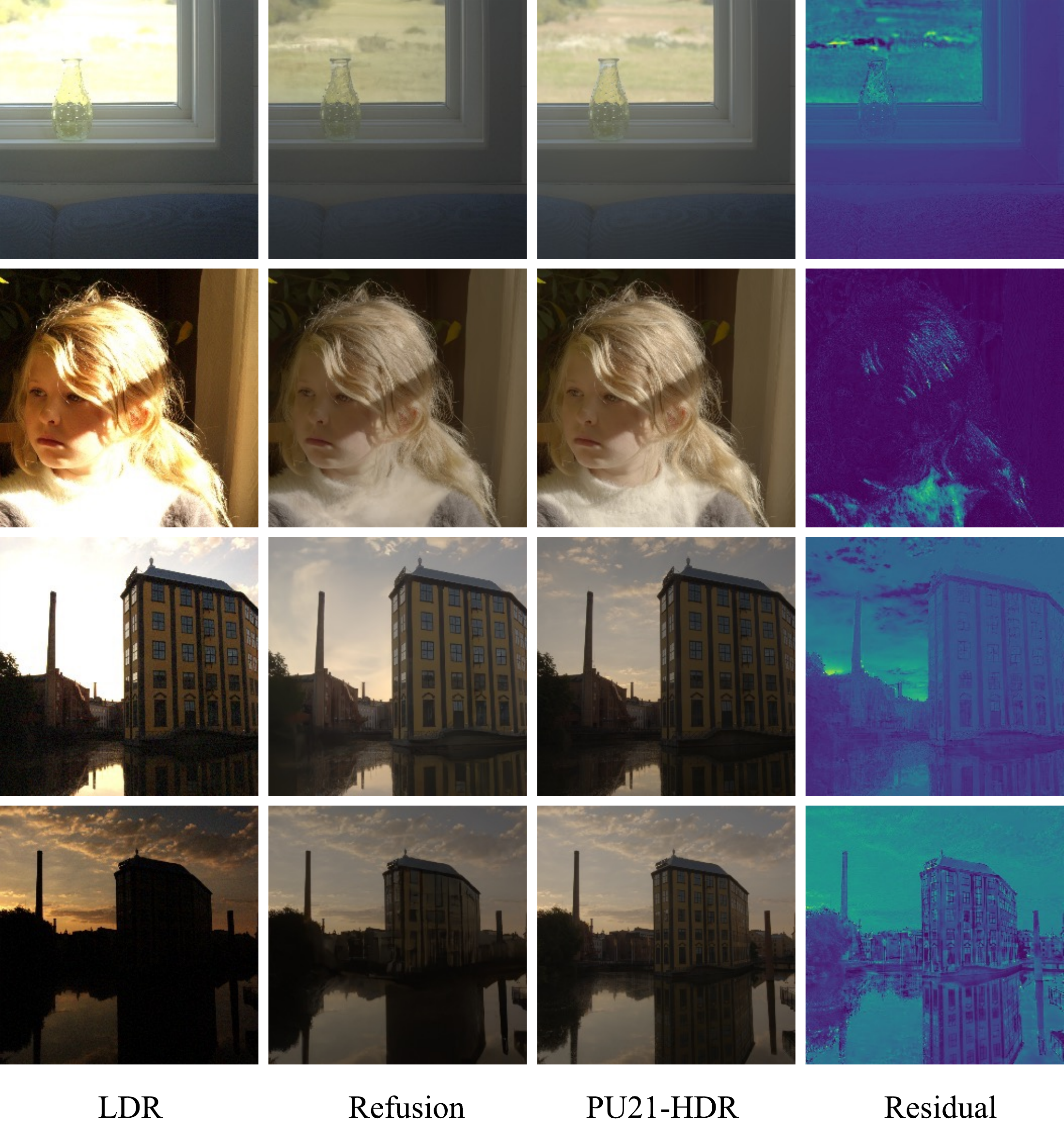}
    \caption{\textit{LiU\_CGIP} error map: Absolute residual between prediction and groundtruth PU21-HDR.}
    \label{team03:fig:errormap}
\end{figure}

\noindent\textit{Qualitative Results}. As shown in Figure~\ref{team03:fig:errormap}, we visualize error maps across different scenarios. Most noticeable errors appear in overexposed regions, particularly in the second and third rows. The last row highlights missing details in underexposed areas, such as completely dark buildings. 

\begin{figure}[!t]
    \centering
    \begin{subfigure}[t]{0.48\linewidth}
        \centering
        \includegraphics[width=\linewidth]{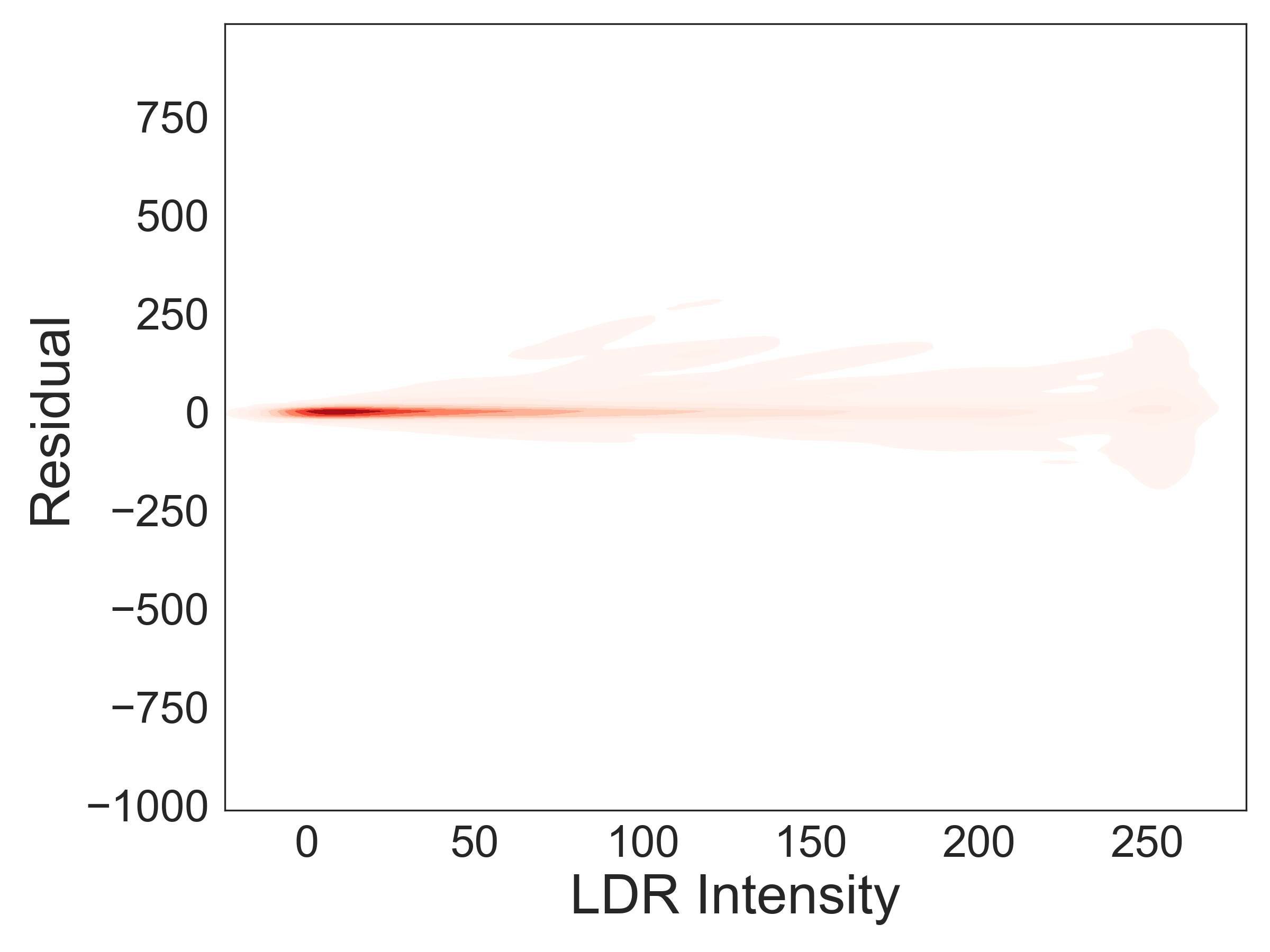}
        \caption{LDR intensity v.s. HDR error}
        \label{team03:fig:kde-all}
    \end{subfigure}
    \hfill
    \begin{subfigure}[t]{0.48\linewidth}
        \centering
        \includegraphics[width=\linewidth]{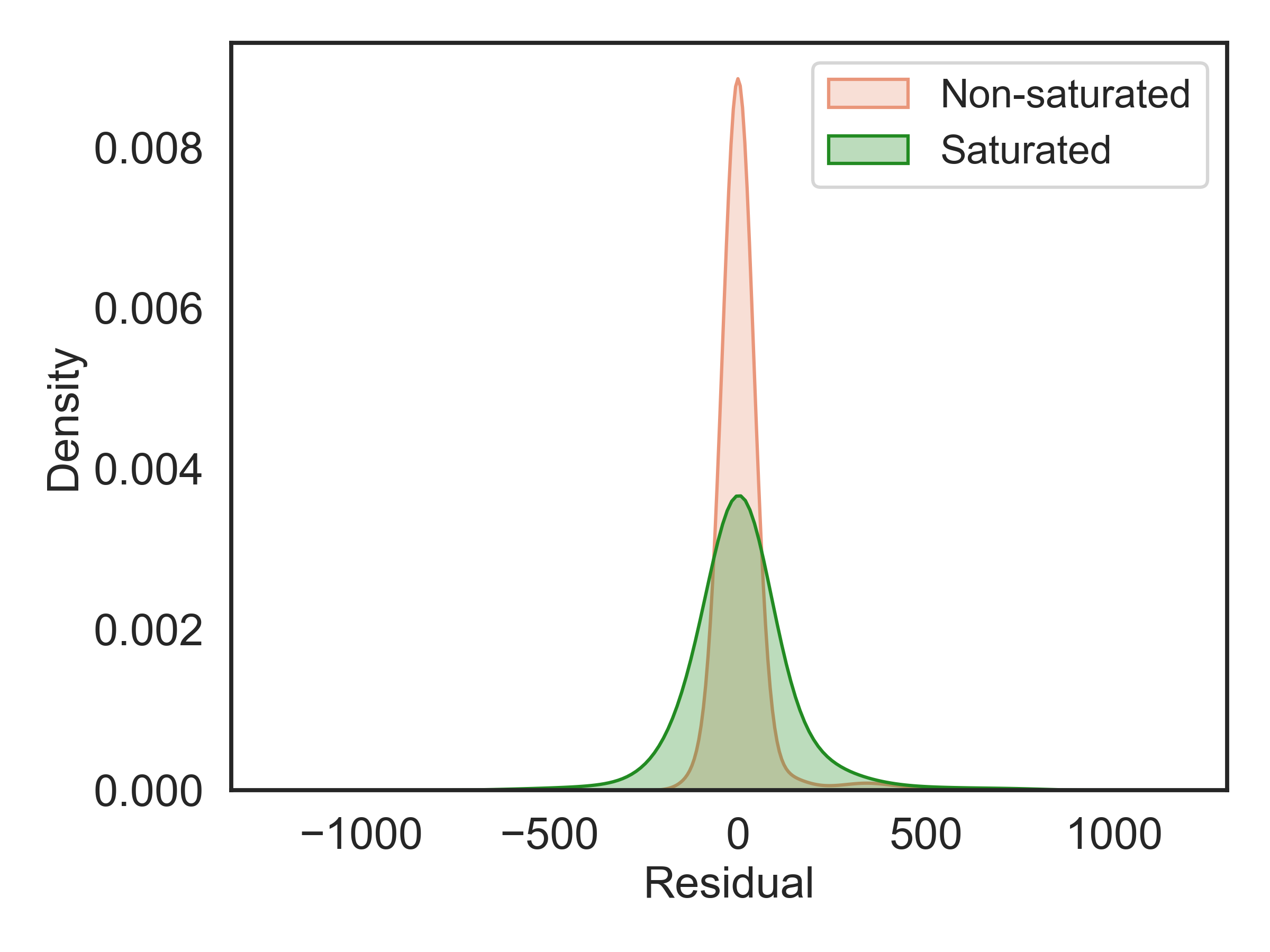}
        \caption{Saturated v.s. non-saturated}
        \label{fig:kde-res}
    \end{subfigure}
    \caption{\textit{LiU\_CGIP}'s error distribution: (a) Joint 2D distribution of input LDR intensity and prediction error. (b) Error distribution on saturated and non-saturated area, \(15\%\) percentile. Both figures are calculated on 100 randomly sampled images from training set.}
    \label{team03:fig:kde-combined}
\end{figure}

To better understand the model behavior, we perform a statistical error analysis. We randomly sample 100 images from the training set and compare LDR intensity values with the corresponding prediction errors. As shown in Figure~\ref{team03:fig:kde-all}, we observe a general trend: prediction error tends to increase slightly with intensity, and a concentration of high errors is visible near intensity value 255.

We define the top 15\% of pixel intensities as the saturated region and analyze their error distribution in Figure~\ref{fig:kde-res}. Compared to the full image distribution, the saturated regions show a distinct pattern with more high-error outliers.

\begin{figure}[!t]
    \centering
    \includegraphics[width=\linewidth]{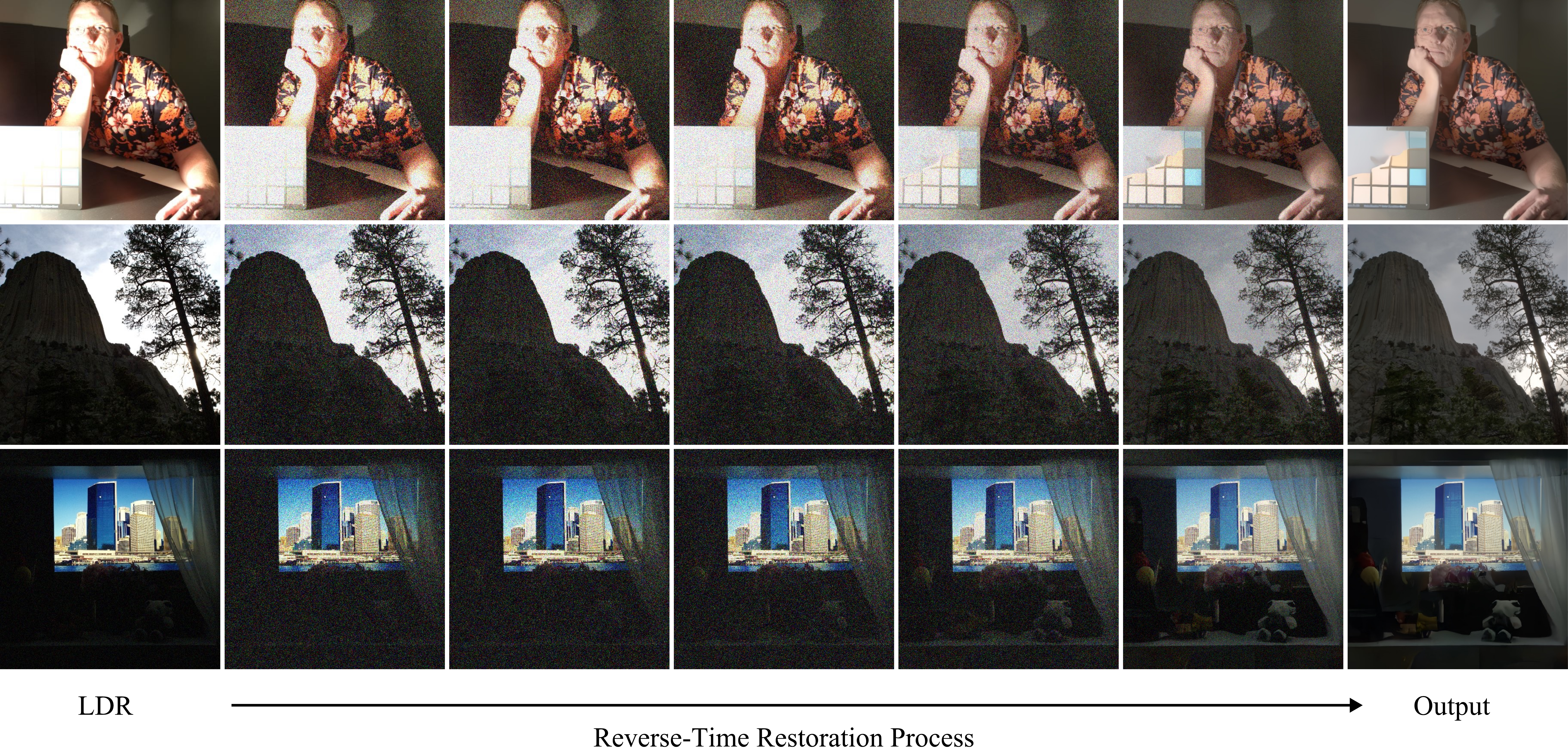}
    \caption{\textit{LiU\_CGIP}'s more results. Samples from development/validation datasets, image id from top to bottom: \texttt{im\_000045\_000002.jpg}, \texttt{im\_000029\_000002.jpg} and \texttt{im\_000057\_000001.jpg}.}
    \label{team03:fig:results}
\end{figure}

Additional visual results are shown in Figure~\ref{team03:fig:results}. For example, in the first row, the reconstruction of an overexposed color chart exhibits unrealistic artifacts. These findings suggest that the most challenging aspect of ITM lies in regions with missing information. Large-scale inpainting models may help address these limitations. Moreover, saturated regions are difficult to evaluate using traditional objective metrics such as PU21-PSNR and PU21-SSIM, which may necessitate the development of more perceptually aligned evaluation methods.

\noindent\textbf{Best scores for development/testing}: 
\begin{itemize}
    \item PU21-PSNR: 34.51/34.33
    \item PU21-SSIM: 0.94/0.94
\end{itemize}

\begin{table}[!t]
    \centering
    \begin{tabular}{lcc}
    \toprule
    \textbf{Method} & \textbf{PU21-PSNR} $\uparrow$ & \textbf{PU21-SSIM} $\uparrow$ \\
    \midrule
    Refusion & 33.62 & 0.9371 \\
    Refusion + extra data & 34.33 & 0.9421 \\
    \bottomrule
    \end{tabular}
    \caption{\textit{LiU\_CGIP}'s ablation on PU21 metrics.}
    \label{team03:tab:pu21-comparison}
\end{table}

\begin{table}[!t]
    \centering
    \resizebox{\columnwidth}{!}{
    \begin{tabular}{l|l|r}
    \toprule
    \textbf{Name} & \textbf{Source} & \textbf{Size} \\ \hline
    EMPA         & \url{http://www.empamedia.ethz.ch/hdrdatabase/index.php}                  & 33             \\ \hline
    HDReye       & \url{http://mmspg.epfl.ch/hdr-eye}                                       & 46             \\ \hline
    Fairchild    & \url{http://rit-mcsl.org/fairchild/HDRPS/HDRthumbs.html}                 & 106            \\ \hline
    Ward         & \url{http://www.anyhere.com/gward/hdrenc/pages/originals.html}          & 33             \\ \hline
    Stanford     & \url{http://scarlet.stanford.edu/~brian/hdr/hdr.html}                   & 91             \\ \hline
    MCSL         & \url{http://www.cis.rit.edu/research/mcsl2/icam/hdr/rit_hdr/}           & 74             \\ \hline
    Funt         & \url{http://www.cs.sfu.ca/~colour/data/funt_hdr/\#DATA}                & 112            \\ \hline
    Boitard      & \url{https://people.irisa.fr/Ronan.Boitard/}                            & 7 sequences    \\ \hline
    MPI          & \url{http://resources.mpi-inf.mpg.de/hdr/video/}                        & 2 sequences    \\ \hline
    DML-HDR      & \url{http://dml.ece.ubc.ca/data/DML-HDR/}                               & 5 sequences    \\ \hline
    HDR book     & Images accompanying the HDR book by Reinhard \textit{et al.} [2005].   & 327            \\ \hline
    JPEG-XT      & Images used in the evaluation by Mantiuk \textit{et al.} [2016].       & 174            \\ \hline
    Stuttgart    & \url{https://hdr-2014.hdm-stuttgart.de/}                                & 33 sequences   \\ \hline
    LiU HDRV     & \url{http://hdrv.org}                                                   & 10 sequences   \\ \hline
    Sequences    & Miscellaneous sequences                                                & 10 sequences   \\ \hline
    Probes       & Miscellaneous lightprobes/panoramas                                    & 12             \\ \hline
    Images       & Miscellaneous images                                                   & 113            \\ \hline
    \textbf{Total}            &    -                                                                    & \textbf{4392 images} \\
    \bottomrule
    \end{tabular}
    }
    \caption{\textit{LiU\_CGIP}'s extra HDR datasets and their sources. We simulated LDR data the same way as in HDRCNN~\cite{EKDMU17}.}
    \label{team03:tab:data}
\end{table}

\subsection{NJ Challenger: Learning to Reverse the Camera Pipeline by PyTorch}

\noindent\textbf{General method description}. Our method is a novel approach for single-image High Dynamic Range (HDR) reconstruction by learning to reverse the camera pipeline. Instead of directly mapping a Low Dynamic Range (LDR) image to an HDR image, the method decomposes the problem into three sub-tasks, each handled by a specialized neural network \cite{liu2020single}:
\begin{itemize}
    \item Dequantization-Net: Restores missing details caused by quantization (e.g., banding artifacts in underexposed regions). The network is U-Net architecture with skip connections to reduce quantization artifacts.

    \item Linearization-Net: Estimates the inverse Camera Response Function (CRF) to convert the non-linear LDR image to a linear irradiance map. This part is ResNet-18 backbone with edge/histogram features and monotonicity constraints for CRF estimation.
    
    \item Hallucination-Net: Predicts missing content in overexposed regions by generating positive residuals. We use Encoder (VGG16) and Decoder with resize-convolution layers to avoid checkerboard artifacts, predicting positive residuals for over-exposed regions.
\end{itemize}
    
The input LDR Image will be processed through three sequential networks. First, the LDR will be inputted into the Dequantization-Net to get a detailed dequantized LDR. Then, Linearization-Net can estimate the inverse, which converts dequantized LDR to linear image. Finally, Hallucination-Net will restore missing content in overexposed regions, and the HDR image will be outputted.

\noindent\textbf{Total method complexity}. The number of parameters of Dequantization-Net is 1.99M. The number of parameters of Hallucination-Net is 24.57M. The number of parameters of Linearization-Net is 1.2M.

\noindent\textbf{Training strategy}. The training is on one 24G NVIDIA GeForce RTX 4060 for a total of 10 days. It takes 3 days to train the Dequantization-Net and also takes 3 days to train the Linearization-Net. And we spend 4 days training the Hallucination-Net. The learning rate scheduler is the cosine annealing scheduler, and the optimizer is the adaptive moment estimation. The learning rate is set to $1\times10^{-4}$. And we train Dequantization-Net for 10K iterations with batch size 64, Linearization-Net for 10K iterations with batch size 64, and Hallucination-Net for 10K iterations with batch size 32. Then we use different losses for each network. 

Dequantization-Net: Trained with l2 loss between dequantized output and ground-truth non-linear image.

Linearization-Net: Uses edge/histogram features and enforces monotonicity constraints. The loss combines l2 for linear image and CRF reconstruction.

Hallucination-Net: Optimized with log-l2 loss (for highlight regions), perceptual loss (VGG features on tonemapped images), and TV loss (for smoothness).

Training data is the official training set: Approximately
19,000 LDR-HDR image pairs at a resolution of $256\times256$, collected from various public HDR datasets. The LDR counterparts are synthesized using a range of
exposure settings, quantization effects, and noise patterns.
We never use any additional training data.

We use pretrained VGG16 and VGG16-places-365 in our Hallucination-Net, VGG16-places-365 is used to initialize the model from the pretrained model, and VGG16 is for perceptual loss. We also use pretrained Crfnet weights to initialize the model for our Linearization-Net.

\noindent\textbf{Experimental results}.
According to official benchmark, the evaluation uses perceptual quality metrics designed for HDR data: 100 LDR inputs will be provided during the final testing phase. The HDR ground truths will remain hidden for unbiased evaluation.

\noindent\textbf{Best scores for development/testing}: 
\begin{itemize}
    \item PU21-PSNR: 29.83 / 29.22
    \item PU21-SSIM: 0.85 / 0.85
\end{itemize}

\subsection{UESTC-ITM: Towards High-Quality Inverse Tone Mapping with Conditional Flow Matching}

\noindent\textbf{General method description}. \textit{Motivation}: Single-image inverse tone mapping (ITM) remains an ill-posed and underconstrained problem, particularly in the presence of saturated highlights, crushed shadows, and quantization artifacts common in legacy LDR imagery. Prior regression-based solutions often fail to hallucinate plausible content in severely clipped or saturated regions, leading to poor detail restoration in reconstructed HDR images. On the other hand, recent advances in generative models have demonstrated strong capabilities in synthesizing perceptually convincing details for various image restoration tasks.

\begin{figure}[!t]
    \centering
    \includegraphics[width=\linewidth]{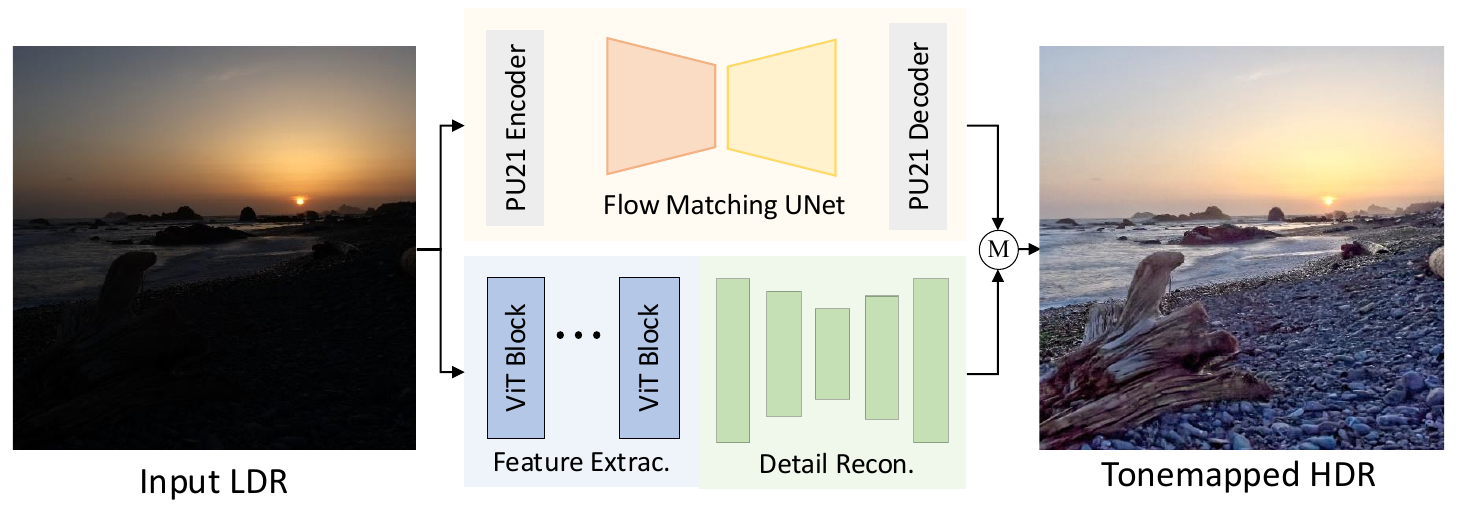}
    \caption{Overview of the \textit{UESTC-ITM}'s proposed ITMFlow. The pipeline comprises two branches: a conditional flow matching (CFM) branch that generates HDR predictions conditioned on the input LDR and noise, enabling robust detail recovery in ill-posed regions; and a ViT+CNN regression branch that ensures high-fidelity reconstruction through global context modeling and local refinement. The final HDR output is obtained by averaging the predictions from both branches.}
    \label{team07:fig:pipeline}
    \vspace{-4mm}
\end{figure}

\noindent\textit{Method Overview}: In this challenge, we propose a dual branch framework, namely ITMFlow, to tackle the inverse tone mapping (ITM) problem using conditional flow matching (CFM)~\cite{tong2023improving}, a principled generative modeling framework that learns to reconstruct HDR images conditioned on the LDR input. By modeling the transformation between LDR and HDR distributions, the CFM branch effectively restores structural and textural details even in highly ill-posed regions. To complement the generative pathway and ensure metric-aligned fidelity, ITMFlow is designed as a dual-branch architecture, where a second branch employs a hybrid CNN + Vision Transformer (ViT) design. Specifically, we adopt NAFNet~\cite{chen2022simple} as a lightweight yet expressive CNN backbone for local detail recovery, while ViT layers~\cite{vision_transformer} capture global contextual information. This deterministic branch is optimized using PU-based metrics to ensure radiometric consistency and quantitative performance. Finally, the outputs of both branches are fused by averaging, effectively combining the perceptual richness of the generative model with the accuracy of deterministic regression. This strategy leads to robust and detail-preserving HDR reconstructions across a wide range of degradations.
The overall pipeline of the proposed ITMFlow is illustrated in Fig.~\ref{team07:fig:pipeline}.
    
\noindent\textbf{Total method complexity}.
\begin{itemize}
    \item Parameters: 69.79M (Flow Matching branch: 29.74M, ViT+CNN branch: 40.05M)
    \item FLOPs: 324.98G (Flow Matching branch: 249.41G, ViT+CNN branch: 75.57G)
    \item Runtime: 1.41s on NVIDIA RTX 4090 GPU for a single image of size 512$\times$512.    
\end{itemize}

\noindent\textbf{Training strategy}. The proposed ITMFlow framework adopts a decoupled training strategy, where the two branches are trained independently. The flow matching branch is trained to reconstruct the PU21-encoded HDR representation. The model learns to predict the conditional velocity field that guides the transformation from noisy LDR inputs to the target PU21-encoded HDR distribution. During inference, the output is passed through a PU21 decoder to obtain the final HDR image. The hybrid ViT+CNN branch is trained in a supervised manner to directly regress the HDR output from the LDR input. At test time, both branches produce independent HDR predictions, which are averaged pixel-wise to obtain the final output. 

Note that we train our model from scratch: we only used the provided data to train and validate our model, and no additional data has been used. 

\begin{figure}[!t]
    \centering
    \includegraphics[width=\linewidth]{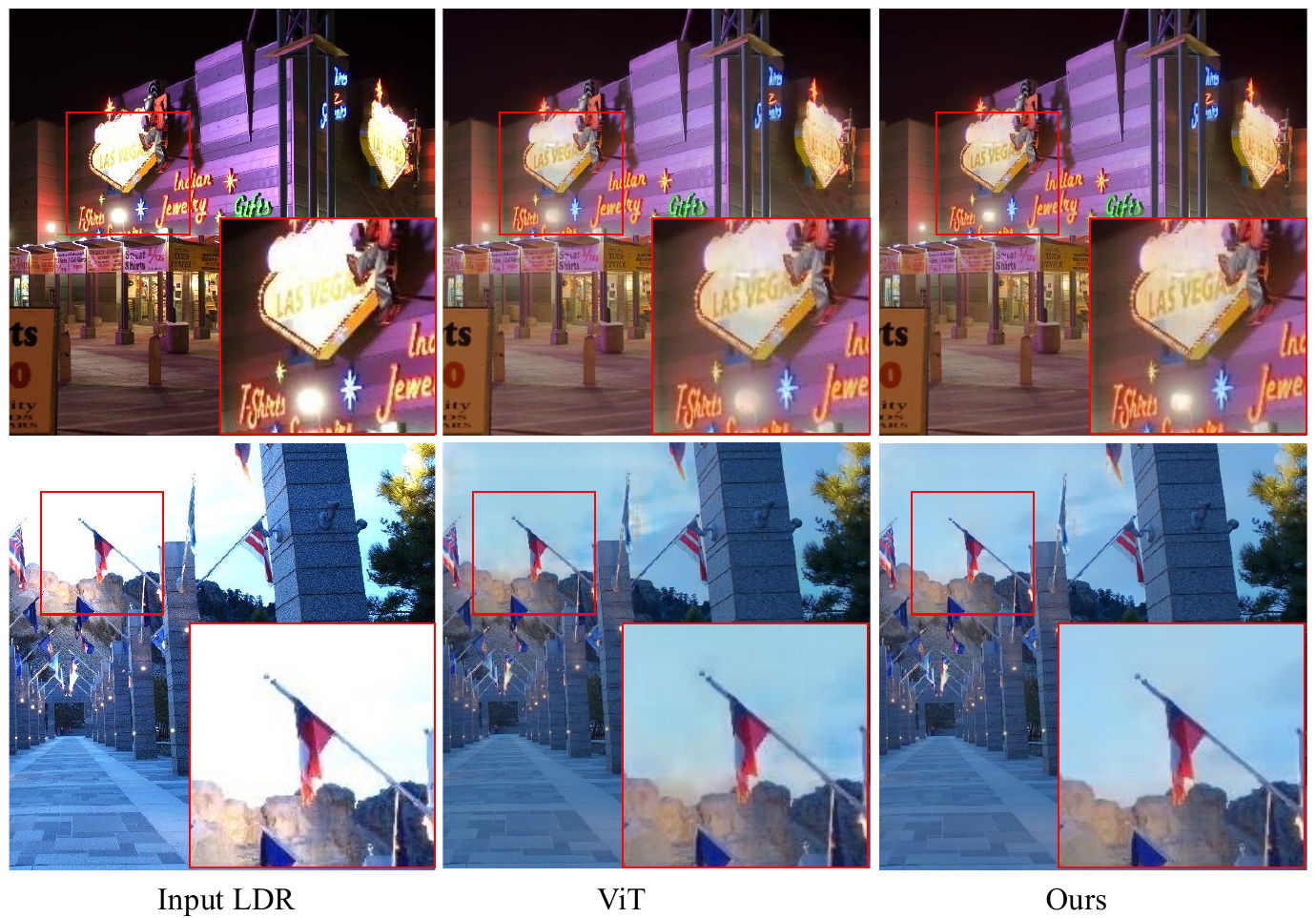}
    \vspace{-2mm}
    \caption{Illustration of the reconstructed HDR results of \textit{UESTC-ITM}.}
    \vspace{-2mm}
    \label{team07:fig:results}
\end{figure}

\noindent\textit{Experimental results}. Figure~\ref{team07:fig:results} illustrates the reconstructed HDR results from ViT and our ITMFlow. Compared to the ViT, ITMFlow achieves better detail reconstruction in overexposed regions and produces results with higher visual consistency and perceptual fidelity.
    
\noindent\textbf{Best scores for development/testing}: 
\begin{itemize}
    \item PU21-PSNR: 33.6 / 34.06 
    \item PU21-SSIM: 0.94 / 0.94
\end{itemize}

\subsection{Jowgik: DITM}

\begin{figure}[!t]
    \centering
    \includegraphics[width=\linewidth]{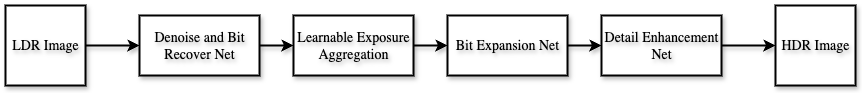}
    \vspace{-4mm}
    \caption{\textit{Jowgik}'s pipeline.}
    \label{team06:fig:pipeline}
    \vspace{-2mm}
\end{figure}

\noindent\textbf{General method description}. DITM is a modular neural architecture for reconstructing HDR images from LDR sRGB inputs, see Figure~\ref{team06:fig:pipeline}. It consists of four main stages: noise-aware bit recovery, learnable exposure aggregation, progressive HDR expansion, and final refinement. The development was influenced by SingleHDR\cite{liu2020single} methodology, which provided valuable insights for our architectural design and loss function formulation.

The main step of the pipeline, given an input $\mathbf{x}_{\text{srgb}} \in \mathbb{R}^{3 \times H \times W}$, are:

\begin{enumerate}
    \item Convert to linear RGB using sRGB function:
    \begin{equation}
        \mathbf{x}_{\text{lin}} =
        \begin{cases}
            \frac{\mathbf{x}_{\text{srgb}}}{12.92}, & \mathbf{x}_{\text{srgb}} \leq 0.04045 \\
            \left(\frac{\mathbf{x}_{\text{srgb}} + 0.055}{1.055}\right)^{2.4}, & \text{otherwise}
        \end{cases}
    \end{equation}
    \item Denoise and recover bit details using \textit{BitRecoverNet}.
    \item Expand dynamic range progressively using \textit{BitExpansionNet}.
    \item Refine spatial details using \textit{DetailEnhancementNet}.
\end{enumerate}

\noindent\textit{BitRecoverNet}. This module removes quantization noise and reconstructs latent HDR content.

Dilated-conv stack estimates a spatial noise map:
\begin{equation}
    \hat{\mathbf{n}} = f_{\text{noise}}(\mathbf{x}_{\text{lin}}) \text{.}
\end{equation}

A residual branch adaptively denoises the signal:
\begin{equation}
    \hat{\mathbf{x}} = \mathbf{x}_{\text{lin}} \cdot \left(1 + \tanh(\beta) \cdot f_{\text{denoise}}([\mathbf{x}_{\text{lin}}, \hat{\mathbf{n}}])\right) \text{.}
\end{equation}

The input is decomposed into under-, mid-, and over-exposed components.

First, blurred luminance is computed as:
\begin{align}
    L &= 0.2126 R + 0.7152 G + 0.0722 B \\
    \bar{L} &= L * k, \quad k=mean\_kernel
    \text{,}
\end{align}

\noindent then, it is softmax-normalized thresholds:
\begin{equation}
    \{\tau_i\}_{i=1}^{2} = \text{cumsum}(\text{softmax}(\theta))
    \text{.}
\end{equation}

Finally, the exposure masks are computed as:
\begin{align}
    M_{\text{under}} &= 1 - \sigma(\alpha (\bar{L} - \tau_1)) \\
    M_{\text{mid}} &= \sigma(\alpha (\bar{L} - \tau_1)) - \sigma(\alpha (\bar{L} - \tau_2)) \\
    M_{\text{over}} &= \sigma(\alpha (\bar{L} - \tau_2)) \text{.}
\end{align}

At this point, weighted fusion of masked images is computed as:
\begin{align}
    \mathbf{x}_{\text{under}} = \hat{\mathbf{x}} \cdot M_{\text{under}} \quad
    \mathbf{x}_{\text{mid}} = \hat{\mathbf{x}} \cdot M_{\text{mid}} \quad
    \mathbf{x}_{\text{over}} = \hat{\mathbf{x}} \cdot M_{\text{over}}
\end{align}

\begin{equation}
    \mathbf{w} = \text{softmax}(f_{\text{w}}([\mathbf{x}_{\text{under}}, \mathbf{x}_{\text{mid}}, \mathbf{x}_{\text{over}}])) \text{.}
\end{equation}

\begin{equation} 
    \mathbf{x}_{\text{fused}} = \sum_{i} \mathbf{w}_i \cdot \mathbf{x}_i \text{.}
\end{equation}

The final enhancement is applied via residual projection:
\begin{equation}
    \mathbf{x}_{\text{deq}} = \mathbf{x}_{\text{fused}} \cdot \left(1 + \sigma(\alpha) \cdot f_{\text{res}}(\mathbf{x}_{\text{fused}})\right) \text{.}
\end{equation}

\noindent \textit{BitExpansionNet} expands the dequantized image to full HDR via a multi-scale encoder-fusion-decoder architecture.

Its encoder has hierarchical features defined as:
\begin{equation}
    x_1 = f_1(\mathbf{x}_{\text{deq}}) \qquad
    x_2 = f_2(x_1) \qquad
    x_3 = f_3(x_2) \text{.}
\end{equation}

Each level uses attention-weighted feature fusion:
\begin{equation}
    A = \sigma(f_{\text{eca}}(\text{avgpool}(x))) \cdot \sigma(f_{\text{spa}}(\text{mean}_c(x)))
\end{equation}
\begin{equation}
    x_{\text{fused}} = f_{\text{conv}}(x) \cdot A \text{.}
\end{equation}

HDR predictions are generated at each scale with skip connections:
\begin{align}
    f_1, o_1 &= \text{proj}_1(x_{\text{fused}}, \text{None}, \mathbf{x}_{\text{deq}}) \\
    f_2, o_2 &= \text{proj}_2(f_1, f_1, o_1) \\
    f_3, o_3 &= \text{proj}_3(f_2, f_2, o_2) \text{.}
\end{align}

\noindent \textit{DetailEnhancementNet}. A shallow U-Net structure further enhances spatial and contrast details of the final HDR prediction.

Skip features are aggregated across five encoder levels and progressively upsampled:
\begin{equation}
    \mathbf{y}_{\text{hdr}} = o_3 \cdot \left(1 + \sigma(\alpha) \cdot f_{\text{enhance}}(o_3)\right)
\end{equation}

\noindent\textbf{Total method complexity}.
\begin{itemize}
    \item Number of parameters: 1.97M
    \item FLOPs: 25.64G (for standard input resolution) 
    \item Number of activations: 85.29M 
    \item Number of Conv2d layers: 81
    \item GPU memory consumption: Efficient processing with adaptive padding for arbitrary resolutions
    \item Runtime: Successfully tested on resolutions including $640\times480$, $321\times241$ with robust handling of arbitrary input dimensions
    \item Model initialization: Kaiming initialization applied to all layers for stable training
\end{itemize}

\noindent\textbf{Training strategy}. During training, the model outputs all intermediate results $[x\_denoised, x\_deq, o_1, o_2, o_3, hdr]$, enabling supervision at multiple stages of the pipeline for improved gradient flow and feature learning. The progressive HDR expansion stages $(o_1, o_2, o_3)$ receive weighted supervision with increasing importance for later stages.

\noindent\textit{Composite Loss Function}. The training employs a sophisticated composite loss combining multiple terms for comprehensive supervision:

Core reconstruction terms:
\begin{itemize}
    \item \textit{Reconstruction Loss}: Weighted $\mu$-law compressed domain L1 loss across all HDR stages, with progressively increasing weights for later stages:
    \begin{equation}
    L_{\text{recon}} = \sum_{i=1}^{N} \frac{i}{N} \cdot ||R_\mu(\text{pred}_i) - R_\mu(\text{gt})||_1
    \text{,}
    \end{equation}
    \noindent where the $\mu$-law range compressor is defined as:
    \begin{equation}
    R_\mu(x) = \frac{\log(1 + \mu \cdot x)}{\log(1 + \mu)}
    \qquad \mu = 5000 \text{.}
    \end{equation}
    \item \textit{Linear Loss}: Direct L1 supervision on final HDR output in linear space:
    \begin{equation}
    L_{\text{linear}} = ||f(\text{input}) - \text{gt}||_1 \text{.}
    \end{equation}
    \item \textit{Denoising Loss}: High-frequency denoising loss using local mean subtraction:
    \begin{equation}
    L_{\text{denoised}} = ||\text{denoised} - \text{gt}||_1 \text{.}
    \end{equation}
\end{itemize}

Perceptual quality terms:
\begin{itemize}
    \item \textbf{$L_{\text{perc}}$}: VGG16-based perceptual loss on tone-mapped outputs using relu1\_2, relu2\_2, relu3\_3, relu4\_3 features.
    \item \textbf{$L_{\text{ssim\_pu}}$}: SSIM loss in Perceptual Uniform (PU) encoding space: $\text{PU}(x) = \frac{\log_{10}(1 + c \cdot x)}{\log_{10}(1 + c)}$ with $c = 10000$ \footnote{From Editors: This is an approximation of PU using the $\mu$-law}.
    \item \textbf{$L_{\text{color}}$}: Log-chrominance consistency using three log-ratio channels (R/G, G/B, B/R) for exposure-invariant color preservation.
\end{itemize}

Advanced quality metrics:
\begin{itemize}
    \item \textit{Unified Patch Fidelity Loss}: Combining focal Charbonnier loss on log-luminance patches, global soft-histogram matching in log domain, and edge-aware smoothness regularization:
    \begin{equation}
    L_{\text{upf}} = L_{\text{charb}} + \alpha_{\text{hist}} L_{\text{hist}} + \beta_{\text{smooth}} L_{\text{smooth}}
    \end{equation}
    \item \textit{Total Variation Loss}: Anisotropic total variation regularization for spatial smoothness:
    \begin{equation}
    L_{\text{tv}} = ||\nabla_h \text{pred}||_1 + ||\nabla_v \text{pred}||_1
    \end{equation}
\end{itemize}

The total composite loss function is:
\begin{equation}
\begin{split}
L_{\text{total}} = &L_{\text{recon}} + \alpha_{\text{perc}} L_{\text{perc}} + \gamma_{\text{ssim}} L_{\text{ssim\_pu}} \\
&+ \gamma_{\text{color}} L_{\text{color}} + \gamma_{\text{tv}} L_{\text{tv}} + \lambda_{\text{linear}} L_{\text{linear}} \\
&+ \alpha_{\text{denoise}} L_{\text{denoised}} + \alpha_{\text{upf}} L_{\text{upf}}
\end{split}
\end{equation}

\noindent Training Configuration:
\begin{itemize}
    \item Input channels: 3 (RGB)
    \item Output channels: 3 (RGB) 
    \item Base feature dimension: 32
    \item Pyramid levels: 3
    \item Loss weights: $\alpha_{\text{perc}} = 0.1$, $\gamma_{\text{ssim}} = 0.1$, $\gamma_{\text{color}} = 0.05$, $\lambda_{\text{linear}} = 0.1$, $\alpha_{\text{denoise}} = 0.1$, $\alpha_{\text{upf}} = 0.1$, $\gamma_{\text{tv}} = 0.1$
    \item UPF parameters: patch size = 16, focal gamma = 1.5, histogram bins = 64, histogram sigma = 0.1
    \item Batch size: 16
    \item Learning rate schedule: Cosine Annealing with base LR set to  $2 \times 10^{-4}$ and annealed to $\text{lr}_{\text{min}} = 1 \times 10^{-6}$, and $T = 350$ epochs
    \item Optimizer: Adam
    \item Training epochs: 350
\end{itemize}

The training pipeline incorporates standard geometric augmentations including horizontal/vertical flipping and rotations to improve model generalization and robustness to different image orientations.

The DITM model is trained from scratch without using any pretrained models (i.e., model trained from random initialization) or external methods using the PyTorch framework with standard components.

\noindent\textbf{Testing strategy}. The model evaluation employs a comprehensive testing framework that supports multiple evaluation modes and datasets. The testing protocol includes:

\noindent \textit{Evaluation Metrics}:
\begin{itemize}
    \item RMSE: Root Mean Square Error in linear HDR domain.
    \item PSNR: Peak Signal-to-Noise Ratio calculated in Perceptual Uniform (PU) encoding space for HDR-appropriate quality assessment.
    \item SSIM: Structural Similarity Index Measure computed in PU domain to better reflect perceptual quality differences.
    \item Runtime: The average inference time per image measured using CUDA events for precise GPU timing.
\end{itemize}

\noindent \textit{Testing Protocol}. The evaluation framework supports arbitrary input resolutions through adaptive padding mechanisms, ensuring consistent processing across different image sizes. The model outputs are evaluated against ground truth in multiple domains:
\begin{itemize}
    \item Linear HDR domain for pixel-level accuracy assessment.
    \item Perceptual Uniform (PU) encoded space using $\text{PU}(x) = \frac{\log_{10}(1 + 10000 \cdot x)}{\log_{10}(1 + 10000)}$ for perceptually-relevant quality metrics \footnote{From Editors: This is an approximation of PU using the $\mu$-law}.
    \item Tone-mapped space using $\mu$-law compression for visualization quality evaluation.
\end{itemize}

\noindent\textbf{Best scores for development/testing}: 
\begin{itemize}
    \item PU21-PSNR: 33.3 / 33.64 
    \item PU21-SSIM: 0.94 / 0.94
\end{itemize}

\section*{Acknowledgments}
This work was partially supported by the Alexander von Humboldt Foundation. We thank the AIM 2025 sponsors: AI Witchlabs and University of W\"urzburg (Computer Vision Lab).

\appendix

\section{Teams and Affiliations}
\label{sec:teams}

\subsection*{AIM 2025 Inverse Tone Mapping Challenge}
\noindent\textit{\textbf{Title: }} AIM 2025 Inverse Tone Mapping Challenge\\
\noindent\textit{\textbf{Members: }} \\
Chao Wang$^{1, 2}$ (\href{winchao1984@gmail.com}{winchao1984@gmail.com}),\\
Francesco Banterle$^3$ (\href{francesco.banterle@isti.cnr.it}{francesco.banterle@isti.cnr.it}),\\
Bin Ren$^{4,5}$ (\href{mailto:bin.ren@unitn.it}{bin.ren@unitn.it}),\\
Radu Timofte $^6$ (\href{mailto:Radu.Timofte@uni-wuerzburg.de}{Radu.Timofte@uni-wuerzburg.de})\\
\noindent\textit{\textbf{Affiliations: }}\\
$^1$ Max-Planck-Institut für Informatik (MPI), Germany\\
$^2$ Peng Cheng Laboratory, China\\
$^3$ ISTI-CNR, Italy\\
$^4$ University of Pisa, Italy\\
$^5$ University of Trento, Italy\\
$^6$ Computer Vision Lab, University of W\"urzburg, Germany\\

\subsection*{ToneMapper}
\noindent\textit{\textbf{Team Name: }} ToneMapper\\
\noindent\textit{\textbf{Title: }} Boosting Inverse Tone Mapping via Regularization Training\\
\noindent\textit{\textbf{Members: }} Xin Lu 
(team leader, \href{mailto:luxion@mail.ustc.edu.cn}{luxion@mail.ustc.edu.cn}), 
Yufeng Peng, 
Chengjie Ge, 
Zhijing Sun, 
Ziang Zhou, 
Zihao Li, 
Zishun Liao, 
Qiyu Kang, 
Xueyang Fu, 
and Zheng-Jun Zha.\\
\noindent\textit{\textbf{Affiliations: }}   University of Science and Technology of China, Hefei, China \\
\noindent\textit{\textbf{Team website: }}\\
\href{https://xueyangfu.github.io/}{https://xueyangfu.github.io/}. 

\subsection*{HDRer}
\noindent\textit{\textbf{Team Name: }} HDRer\\
\noindent\textit{\textbf{Title: }} Deep High Dynamic Range Imaging via Dynamic Scenes Generation\\
\noindent\textit{\textbf{Members: }} Xin Lu 
(team leader, \href{mailto:luxion@mail.ustc.edu.cn}{luxion@mail.ustc.edu.cn}), Zhijing Sun, Chengjie Ge, Xingbo Wang, Kean Liu, Senyan Xu, Yang Qiu, Qiyu Kang, Xueyang Fu, and Zheng-Jun Zha.\\
\noindent\textit{\textbf{Affiliations: }}   University of Science and Technology of China, Hefei, China \\
\noindent\textit{\textbf{Team website: }}\\
\href{https://xueyangfu.github.io/}{https://xueyangfu.github.io/}

\subsection*{LiU\_CGIP}
\noindent\textit{\textbf{Team Name: }} LiU\_CGIP\\
\noindent\textit{\textbf{Title: }} What Makes Inverse Tone Mapping Hard? An Empirical Study of Error Patterns and Perceptual Challenges\\
\noindent\textit{\textbf{Members: }} Yifan Ding
(team leader, \href{mailto:yifan.ding@liu.se}{yifan.ding@liu.se}), Gabriel Eilertsen, and Jonas Unger.\\
\noindent\textit{\textbf{Affiliations: }}  Linköping University, Sweden \\
\noindent\textit{\textbf{Team website: }}\\
\href{https://github.com/limchaos/Refusion-HDR.git}{https://github.com/limchaos/Refusion-HDR.git}. 

\subsection*{NJ Challenger}
\noindent\textit{\textbf{Team Name: }} NJ Challenger\\
\noindent\textit{\textbf{Title: }} Learning to Reverse the Camera Pipeline by PyTorch\\
\noindent\textit{\textbf{Members: }} Zihao Wang
(team leader, \href{wzh960045@outlook.com}{wzh960045@outlook.com}), Ke Wu, and Jinshan Pan.\\
\noindent\textit{\textbf{Affiliations: }} Nanjing University of Science and
Technology, China\\
\noindent\textit{\textbf{Team website: }}\\
\href{https://github.com/wzh960045/singhdr-torch}{https://github.com/wzh960045/singhdr-torch}. 

\subsection*{UESTC-ITM}
\noindent\textit{\textbf{Team Name: }} UESTC-ITM\\
\noindent\textit{\textbf{Title: }} Towards High-Quality Inverse Tone Mapping with Conditional Flow Matching\\
\noindent\textit{\textbf{Members: }} Zhen Liu
(team leader, \href{mailto:liuzhen03@std.uestc.edu.cn}{liuzhen03@std.uestc.edu.cn}), Zhongyang Li, and Shuaicheng Liu.\\
\noindent\textit{\textbf{Affiliations: }} University of Electronic Science and Technology of China (UESTC), China\\
\noindent\textit{\textbf{Team website: }} None. 

\subsection*{Jowgik}
\noindent\textit{\textbf{Team Name: }} Jowgik\\
\noindent\textit{\textbf{Title: }} DITM\\
\noindent\textit{\textbf{Members: }} S. M. Nadim Uddin 
(team leader, \href{mailto:smnadimuddin@gmail.com}{smnadimuddin@gmail.com}).\\
\noindent\textit{\textbf{Affiliations: }} Deep In Sight Co., South Korea \\
\noindent\textit{\textbf{Team website: }}\\
\href{https://github.com/SayedNadim/ITM25}{https://github.com/SayedNadim/ITM25}.

{
    \small
    \bibliographystyle{ieeenat_fullname}
    \bibliography{ref}
}

\end{document}